\documentclass[letterpaper, 10 pt, conference]{ieeeconf}  
\IEEEoverridecommandlockouts      
\usepackage[pdftex]{graphicx}
\usepackage{listings}
\usepackage{amsmath,amssymb,amsfonts} 
\usepackage{algorithm,algorithmic}
\usepackage{color}
\usepackage{lipsum}
\usepackage{caption}
\usepackage{comment}
\usepackage{subcaption}
\usepackage{url}            
\usepackage{booktabs}      
\usepackage{algorithm,algorithmic} 
 
\DeclareMathOperator*{\argmax}{argmax}

\newcommand\Tau{\mathcal{T}}
\usepackage{tabularx}
\usepackage{cite}
\usepackage{hyperref}
\usepackage[absolute,overlay]{textpos}

\title{\LARGE \bf
Learning Modular Robot Locomotion from Demonstrations
}
\author{Julian Whitman and Howie Choset
\thanks{Carnegie Mellon University, Pittsburgh PA, USA.
        {\tt\small jwhitman@cmu.edu}}%
}

\begin{document}

\maketitle
\thispagestyle{plain}
\pagestyle{plain}

\begin{textblock*}{5in}(0.55 in,0.5in) 
Preprint: Submitted to the 2023 International Conference on Robotics and Automation
\end{textblock*}

\begin{abstract}
Modular robots can be reconfigured to create a variety of designs from a small set of components. But constructing a robot's hardware on its own is not enough-- each robot needs a controller. One could create controllers for some designs individually, but developing policies for additional designs can be time consuming. This work presents a method that uses demonstrations from one set of designs to accelerate policy learning for additional  designs. We leverage a learning framework in which a graph neural network is made up of modular components, each component corresponds to a type of module (e.g., a leg, wheel, or body) and these components can be recombined to learn from multiple designs at once. In this paper we develop a combined reinforcement and imitation learning algorithm. Our method is novel because the policy is optimized to both maximize a reward for one design, and simultaneously imitate demonstrations from different designs, within one objective function. We show that when the modular policy is optimized with this combined objective, demonstrations from one set of designs influence how the policy behaves on a different design, decreasing the number of training iterations needed.

\end{abstract}

\section{Introduction} \label{sec:intro}

This paper introduces a framework for combination reinforcement and imitation learning (RL+IL) to create policies for modular robots. 
Modular robots allow a variety of designs to be constructed from a small set of inter-changeable hardware components.
But, constructing the robot is only the first step toward deploying it. 
Each design needs a control policy, and it can be time-consuming to develop a new policy for each new robot, especially given the large number of the designs one could create.

We address this problem by training new robots to exhibit behaviors demonstrated by different existing robots. 
That is, the novel feature of our method is that given demonstrations from multiple robots with different designs, we can train a policy to control a robot with yet a different design from those shown in the demonstrations.
This is where our RL+IL work differs from prior efforts-- related work assumes the robot in demonstrations and the robot trained have the same designs \cite{RoboImitationPeng20, hester2018deep, ding2019goal, xie2020learning}, for instance, using demonstrations from a robot with four legs to train a robot with four legs. 
We constrain our methods to robots composed from combinations of a set of modules, but allow robots to have varying numbers and types of limbs.

\begin{figure}[tb] 
     \centering
      \includegraphics[width=.98\linewidth]{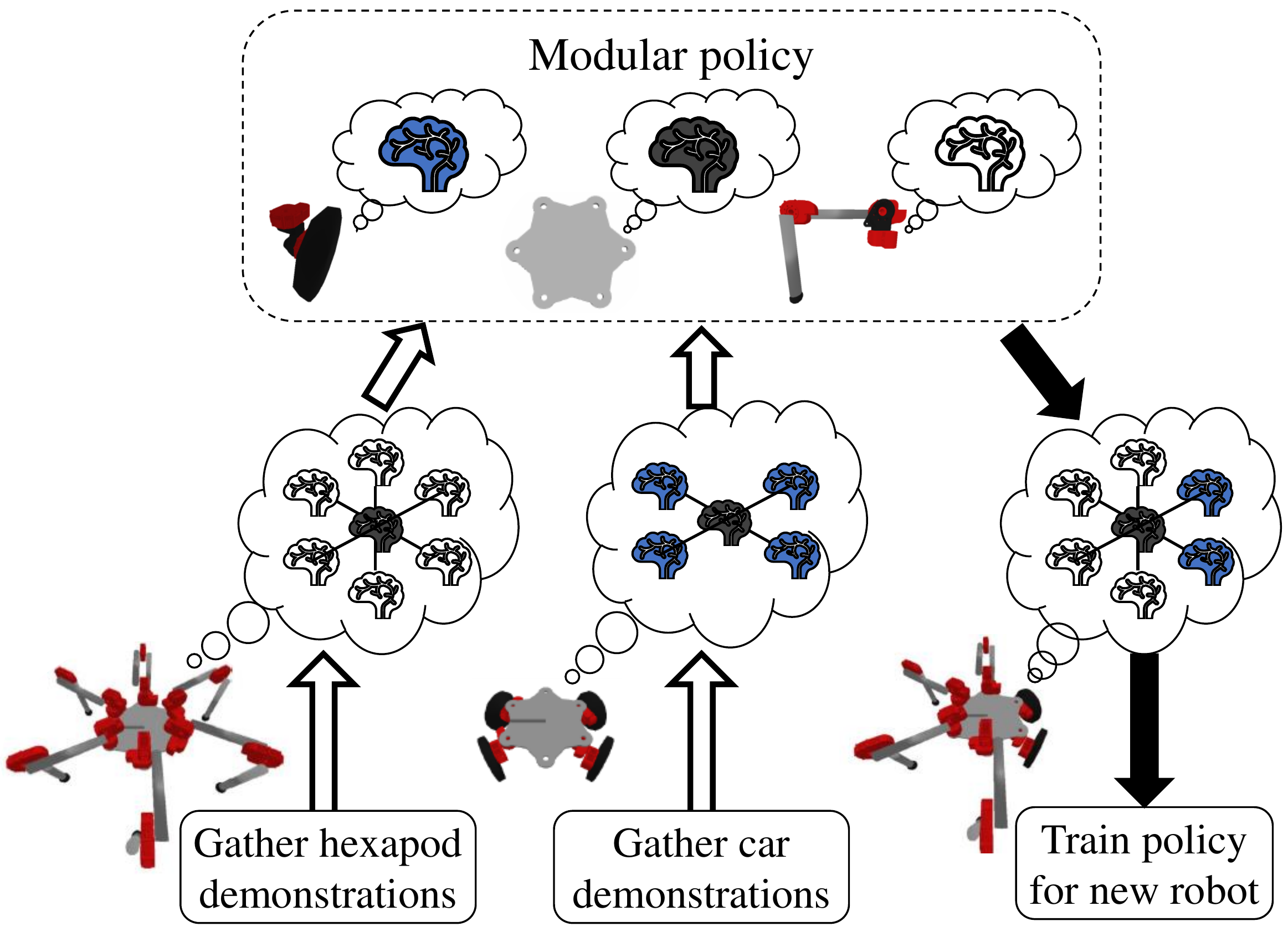}
      \caption{
        This work presents a new combination reinforcement and imitation learning algorithm for modular robots.
        With our method, demonstrations can come from robots with multiple different designs, and can be different yet than the design the policy is trained to control.
        A modular policy (top) consists of neural network components used by each module, represented by brain icons. All modules of a given type use the same neural network, e.g., all wheels use the same blue ``brain'' even when they are placed in different locations on a single robot or placed in different robots. 
        The policy can be applied to different designs by connecting the components into a graph, where the graph structure corresponds to the connectivity of modules in the robot. 
        In this example, data from a hexapod (bottom left) and car-like robot (bottom center) are used to train a robot with both legs and wheels (bottom right). 
        The demonstrations are used in the imitation learning part of training loss indicated by the unfilled arrows, and the leg-wheel robot is used in the reinforcement learning part of the loss, indicated by the filled arrow. 
      }
      \label{fig:modular_policy_rl_il}
\end{figure}

The RL+IL method introduced here, as its name suggests, optimizes the policy by summing two parts: a reinforcement learning (RL) part that maximizes the reward received from robot-environment interactions, and an imitation learning (IL) part that minimizes the difference between the actions output by the policy and the actions shown through demonstrations of a system executing a task. 
The demonstrations can be represented as a dataset of trajectories, which do not have an associated reward, and in fact, we may not even know how ``good'' the demonstrations are. 
We leverage the demonstrations to bootstrap RL for new designs, beyond those designs present in the demonstrations, such that the demonstrations act effectively as a ``prior'' for the policy.

The ability to ``mix and match'' between the design that the policy is trained to control (for the RL part) and the designs in the demonstrations (for the IL part) stems from the fact that we use \textit{modular policies}.
Our prior work \cite{whitman2021learning} introduced the concept of  modular policies\footnote{Our use of the term ``modular polices'' shares similarities with the ``shared modular policies'' of \cite{huang2020smp}.} whose intrinsic structure allows them to train with multiple robot designs.
In the modular policy framework, the policy is divided up into inter-changable components, each of which corresponds to a type of module (e.g., a leg, a wheel, or a body) (Fig. \ref{fig:modular_policy_rl_il}).
To apply the policy to a robot, first the robot is represented as a graph, where modules represent nodes and edges represent electromechanical connections between them.  
Then, the components of the modular policy are reconfigured into a graph matching the structure of the physical robot. 
By reconfiguring the policy during training, it can learn from and apply to a variety of designs at once.
Previously, modular policies were trained with RL. 
We develop a RL+IL method with modular policies, where the policy is reconfigured to apply to different designs in the two parts of the training objective function.

We apply our method to train a locomotion policy for robots with different combinations of legs and wheels.
Our experiments show how including demonstrations, even from robots with different designs than the robot being trained, can accelerate learning, by causing the policy to converge in fewer iterations than when learning without demonstrations.

\section{Background}
\subsection{Reinforcement and imitation learning}

RL can train a control policy, often implemented as a deep neural network, using data collected iteratively from an agent's interaction with its environment \cite{ibarz2021train}.
RL methods have been developed recently to control a variety of robots including articulated locomoting robots \cite{ibarz2021train,  heess2017emergence, hafner2020towards, zhang2021importance, amos2021model, schulman2017proximal }.
However, the most frequently-used RL formulations assume that robot behavior must be discovered without prior knowledge.

Prior knowledge can be inserted into RL methods most often through one of two approaches.
Firstly, a specific motion can be encouraged via reward shaping, e.g., rewarding a specific footfall pattern \cite{whitman2021learning, rudin2022learning}.
Secondly, motions can be forced by constructing action spaces which superpose policy outputs onto a hand-crafted periodic pattern, e.g., perturbing a nominal trot gait output by an open-loop trajectory generator \cite{simtoreal2018, iscen2018policies, yang2020data, hwangbo2019learning, lee2020learning }.
Both of these methods require experts to manually identify and specify desired motions for the robots being trained, and alter the rewards or actions accordingly.

A policy can also be taught from prior knowledge using IL.  
In IL, the policy is trained to reproduce the actions taken within a dataset of demonstrations.
For example, the most long-standing IL method known as ``behavioral cloning'' trains a policy with supervised regression \cite{pomerleau1988alvinn, kaufmann2020deep}.
Expert demonstrations typically come either from humans \cite{ross2011reduction, hawke2020urban} or a model-predictive controller \cite{pan2018agile, pan2020imitation}. 
When the demonstrations are not sufficient on their own to learn a global policy, RL+IL can be combined to leverage a dataset of demonstrations without being limited to states and actions contained therein \cite{hester2018deep, ding2019goal, RoboImitationPeng20, xie2020learning, sartoretti2019primal}.
For example, \cite{hester2018deep} combined RL and IL by training a value function approximation network from state transitions in a dataset.  
\cite{ding2019goal} used expert demonstrations within a variant of hindsight experience replay \cite{andrychowicz2017hindsight}.
\cite{RoboImitationPeng20} constrained the states visited by the policy to be near the states in the dataset.
A limitation of these methods is that they require the robot design in the demonstrations be similar to the robot being trained, for instance, having the same number and type of joints and limbs.

\subsection{Modular policies} \label{sec:modular_policies}

Our past work \cite{whitman2021learning} introduced the \textit{modular policy} framework, geared toward systems where multiple designs are constructed from a set of modules. 
Modular policies leverage the fact that the structure of a modular robot can be represented as a design graph, with nodes as modules and edges as connections between them. 
A design graph is used to create a policy graph with the same structure (Fig. \ref{fig:modular_policy_rl_il}). 
Within a policy graph, each type of module has a neural network associated with it-- i.e., there is one network used to control all of the ``leg'' modules, and a different one used to control all of the ``wheel'' modules.
Each node in the policy graph uses the neural network associated with its corresponding module type to compute a module's actions from its observations. 
For example, a single neural network for leg-type modules is used to compute actions for each of the legs on a hexapod, one leg at a time.

Modular policies are implemented using message-passing graph neural networks (GNN) \cite{scarselli2008graph,wang2018nervenet}, which allows modules to learn to modulate their outputs via a communication procedure in which they exchange information over the graph edges. 
As a result, the policy produces different behaviors for a module depending on its location within the robot, and the presence of other modules within the robot.
In this work, we use the policy architecture of \cite{whitman2021learning}, but train it with a new RL+IL algorithm.

\section{RL+IL for modular robots} \label{sec:mbrl_il}

We are not the first to use RL \cite{whitman2021learning, nagabandi2018neural, janner2019trust, yang2020data} nor RL+IL \cite{RoboImitationPeng20, hester2018deep, ding2019goal} for robot locomotion, and as such our methods use concepts from these prior methods.
The main novelty of our method lies in the combination of modular policies with an RL+IL algorithm.
Recall our prior work \cite{whitman2021learning} used an RL approach to train one modular policy for multiple designs at once. 
In this work, we add an IL part to the training, which influences the behaviors the policy learns with the RL part, such  that a new robot can learn from behaviors demonstrated by other robots with different designs.
We next pose the policy optimization problem addressed, then describe how RL and IL objectives are formulated, followed by training algorithm details.

\begin{figure}[tb] 
\vspace{0.5em}
     \centering
      \includegraphics[width=0.98\linewidth]{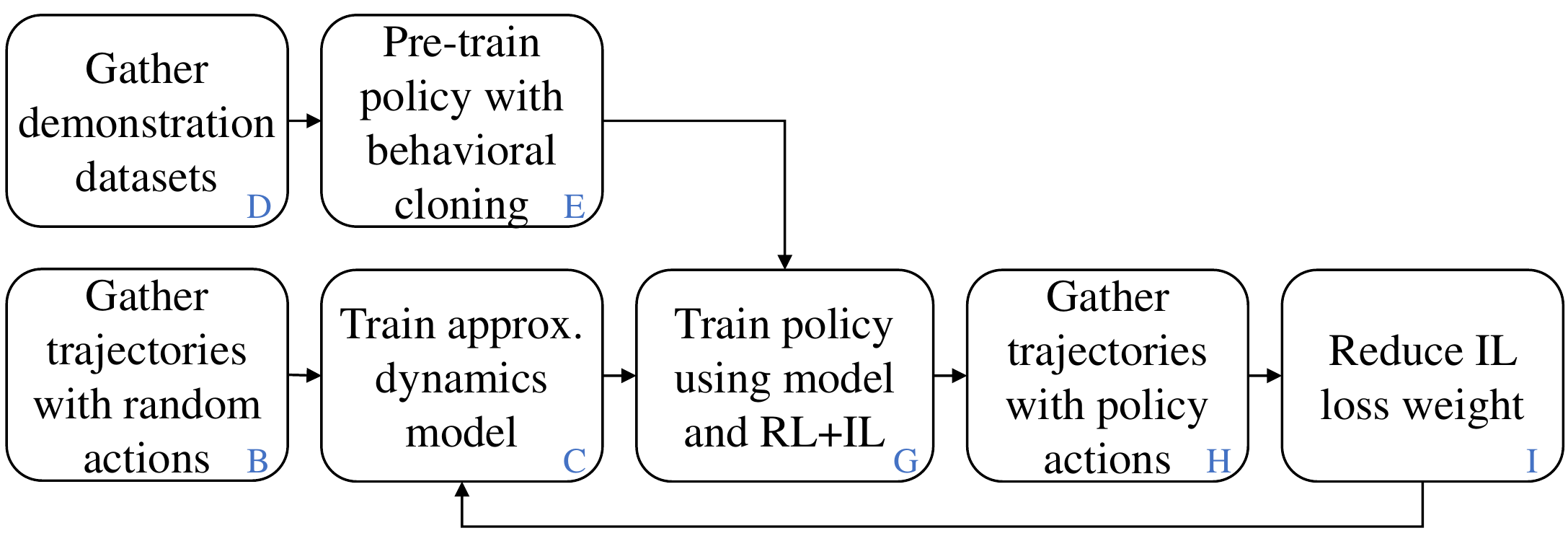}
      \caption{A flow chart depicting the RL+IL algorithm phases. Each block in this diagram is described in Sec. \ref{sec:mbrl_il}, where the subsection label is shown in blue in the corner of each block. 
      }
      \label{fig:pipeline_il}
\end{figure} 

\subsection{Problem formulation}

The design $d_\textrm{train}$, which the policy is being trained to control is first chosen by the user.
The objective we optimize, as is conventionally used in deep RL, maximizes the reward a policy would accumulate when applied to that design.
This maximization problem is,
\begin{align} \label{eqn:policy_opt}
     \theta^* =  &\argmax_{\theta} \mathbb{E}_{a \sim \pi_\theta} \sum_{t=1}^T r (s_t, a_t, d_\textrm{train} )  \\
    & \textrm{s.t.} \quad s_{t+1} = f(s_t, a_t, d_\textrm{train})  \qquad \textrm{(Dynamics)}\\
   & a \sim \pi_\theta(a|o,  d) \qquad \textrm{(Policy)} \\
   & o \sim O(o|s) \qquad  \textrm{(Observation function)} \\
   & s_o \sim p(s) \qquad  \textrm{(Initial state distribution.)} 
\end{align}
The policy $\pi_\theta$ is a distribution over actions, with parameters $\theta$, conditioned on a design and observations. 
The reward $r:S\times A \times \mathcal{D}  \to  \mathbb{R}$ is summed over a time horizon $T$.
The observation function $O$ adds noise to, or removes, parts of the state vector. 
We choose the same observation function as \cite{whitman2021modular}, which removes the x/y position, yaw, and the linear velocity of the body from the state, with Gaussian noise added to the remaining joint and IMU sensor readings.

We solve (\ref{eqn:policy_opt}) by developing a model-based reinforcement learning (MBRL) method. 
We choose this class of policy optimizer over others possible, such as model-free reinforcement learning (MFRL), primarily because MBRL has been shown by past studies to be more data-efficient than MFRL \cite{chua2018deep,amos2018differentiable,janner2019trust, rajeswaran2020game}. 
An MBRL algorithm learns an approximate model $\tilde f_\phi$ with  parameters $\phi$ and then optimizes the policy parameters with that model as a differentiable stand-in for the true dynamics.    
Our algorithm is inspired in particular by the algorithms of \cite{whitman2021learning} and \cite{janner2019trust}, and similarly iterate between collecting data, training the model, and training the policy.
The phases of the algorithm, which are depicted in flow-chart form in Fig. \ref{fig:pipeline_il}, and psuedocode in Algorithm \ref{algo:mbrl_il}.

\subsection{Gather trajectories with random actions}
\label{sec:initial_data}
First, the robot $d_\text{train}$ is randomly initialized at states perturbed from its nominal position.
We create random actions by sampling $10$ values from a normal distribution for each degree-of-freedom, and fitting splines to create smooth joint commands over $100$ time steps, similar to the methods used by \cite{nagabandi2018neural, whitman2021learning}.
These actions are applied to obtain an initial dataset of trajectories $\mathcal{T}_\textrm{train}$.

\subsection{Train approximate dynamics model} \label{sec:model_learning}

Next, the dataset $\mathcal{T}_\textrm{train}$ is used to fit an approximate dynamics model for $d_\textrm{train}$, with the same supervised model learning methods described by \cite{whitman2021learning}. 
The model $\tilde f_\phi$ is a modular model with the same structure as the modular policy, i.e., $\tilde f_\phi$ is implemented as a graph neural network with body, leg, and wheel module node types.
The body node takes as input the state of the body (world frame orientation and velocity), and the leg and wheel nodes take in the joint angles and joint velocities from their respective modules.
The model outputs an estimated change in state given a state and action input.

\subsection{Gather demonstration datasets}
The set $D_{\text{demo}}$ can contain multiple designs, each of which has multiple trajectories that form a demonstration dataset. 
We define a demonstration trajectory $\tau$ as a sequence of observations and actions for a single design over an $H$-step horizon.
Each design has a set of $L$ trajectories, $\Tau^i$, for a design $i \in \{1 \dots |D_{\text{demo}}|\}$.
The collection of these trajectories make up the full demonstration dataset $\Tau_{\text{demo}}$,
\begin{equation}
    \begin{split}
\tau = \{(o_0, a_0), \dots (o_H, a_H)\}, \quad \Tau^i = \{\tau^i_1 \dots \tau^i_L \}, \\
\Tau_{\text{demo}} = \{\Tau^1, \dots \Tau^{|D_{\text{demo}}|} \} 
\end{split}
\end{equation}
These demonstrations do not need to contain full states nor rewards, and can be created via an existing policy such as open-loop hand-crafted behaviors, or human tele-operation.

\subsection{Pre-train policy with behavioral cloning}

Before the RL phase of policy optimization begins, we use IL to ``pre-train'' a policy, that is, to warm-start the policy for the later RL+IL phase. 
In policy pre-training, we employ behavioral cloning using the demonstration dataset.

To encourage the policy to imitate the actions in the dataset we use a log likelihood loss,
\begin{equation}  \label{eqn:IL_loss} 
     \mathcal{L}_{\textrm{IL}} = - \sum_{d\in D_{\textrm{demo}}}
     \sum_{(o,a)  \in \Tau_{\textrm{demo}}}     \log( \pi_\theta(a | o, d)  ). 
\end{equation}
This IL loss is the joint log likelihood that the policy will produce the actions from the dataset when given the observations from the dataset. 
This loss is different from a conventional behavioral cloning loss because it includes a summation over designs.
Note each design is made up from a different combination of modules, each of which may have a different number of degrees-of-freedom (and thus a different action and observation dimensionality).
We can compute this loss using a modular policy because the modular policy can apply to multiple designs.

The policy trained from this loss is not effective globally-- it does not know how to recover to from arbitrary random positions drawn from an initial state distribution $p(s)$. 
It also only uses $D_\textrm{demo}$, and is not applied at this phase to $d_\textrm{train}$.
However, it serves to initialize the policy for further training. 
This phase can also be conducted in parallel with model learning (Sec. \ref{sec:model_learning}).

\subsection{Train policy using model and RL}

The policy is optimized to minimize objective cost (maximize reward) with respect to the model.
In this phase, model parameters are frozen, and the policy is used to compute a series of states, actions, and observations.
The RL loss function derived from (\ref{eqn:policy_opt}) is
\begin{equation} \label{eqn:RL_loss}
     \mathcal{L}_{\textrm{RL}} =  - \mathbb{E}_{a \sim \pi_\theta} \sum_{t=1}^T r (\tilde s_t, a_t,  d_\textrm{train}).  
\end{equation}
The policy parameters are optimized with respect to roll-outs with the approximate dynamics model, where the approximate state evolves according to the model $\tilde s_{t+1} = \tilde f_\phi(\tilde s_t, a_t, d_\textrm{train})$.
The initial state $\tilde s_t = s_0$ is taken from the simulation.
Actions are sampled according to the policy, $a_t \sim  \pi_\theta(a_t|o_t, d_\textrm{train})$, and observations from the observation function $o_t \sim O(\tilde s_t)$.

Over the time horizon length $T$, the policy is applied to the approximate state, a reward is computed, and the state is advanced according to the model.
The total reward is accumulated in $\mathcal{L}_{\textrm{RL}}$. 
After a $T$-step roll-out of forward passes using the model and policy, the policy parameter gradients can be computed by back-propagating through the end-to-end differentiable sequence of states and actions.
These forward and backward passes are repeated for $K$ steps with random mini-batches of $N_\textrm{batch}$ initial states. 
This process takes advantage of the fact that the a neural network-based model is differentiable and can be applied in large parallel batches.

The main term in the reward function $r$ rewards the locomotion forward in the $+x$ direction.
Other terms with smaller relative weights penalize roll, pitch, yaw, deviation from $y=0$, control effort, and distance from a nominal pose. 
The reward does not include motion style terms, and the action space is commanded joint velocities without the need for any open-loop user-defined trajectory.

\subsection{Train policy using model and RL+IL}  \label{sec:il}

On its own, (\ref{eqn:policy_opt}) searches for a policy without prior knowledge.
We next turn to incorporating behavioral priors derived from a set of trajectories from designs $D_\text{demo}$.

To optimize the policy parameters so that they achieve both the RL and IL objectives, we can add the IL loss (\ref{eqn:IL_loss}) to the RL loss (\ref{eqn:RL_loss}) with weight $\lambda \in \mathbb{R}^+$, 
\begin{align} 
     \mathcal{L} &= \mathcal{L}_{\textrm{RL}} + \lambda \mathcal{L}_{\textrm{IL}}. \label{eqn:RLIL_loss} 
\end{align}
and optimize policy parameters with respect to this combined loss, with stochastic gradient descent on $d\mathcal{L}/d\theta$. 

Where in the earlier pre-training phase, the policy was optimized only with respect to the IL loss, in this phase the policy is optimized with respect to the combined RL and IL loss (\ref{eqn:RLIL_loss}).
At each iteration in the policy optimization phase, a batch of trajectories are sampled from $\mathcal{T}_\textrm{demo}$, used to compute the IL loss, and added to the RL loss. 
As a result, the policy is encouraged to reproduce the actions in the demonstrations, without being limited to those behaviors.
With this loss, the policy can learn recovery behaviors and proceed from initial states outside of those contained in the demonstration dataset.

A novel aspect of our RL+IL method is that \textit{demonstration designs in $\mathcal{L}_{\textrm{IL}}$ do not need to be the same as the training designs in $\mathcal{L}_{\textrm{RL}}$}. 
This is possible because the modular policy enables one set of neural network parameters $\theta$ to apply to multiple different designs.
As long as $d_{\text{train}}$ and designs in $D_{\text{demo}}$ are constructed with combinations of modules, our RL+IL method allows for robots with different designs to learn from one another.

\subsection{Gather trajectories with policy actions}

After the policy training phase, the policy is used to gather additional trajectories in simulation.
Policy optimization can exploit model bias \cite{deisenroth2011pilco} to produce a policy that will be low-cost under the model but could be high-cost in the simulation environment.
To correct model bias near states created by the policy, we apply the policy to gather trajectories, and use those trajectories to fine-tune the model.
At the next iteration of policy optimization the model will be more accurate, causing a virtuous cycle in which the model improves near states visited by the policy, and the policy is optimized with respect to the refined model.
The policy is applied without added noise to the simulation environment. 
Then, it is applied with time-correlated noise where the noise is created in the same manner as in Sec. \ref{sec:initial_data}, but with a lower noise variance. 
The resulting trajectories are added to $\mathcal{T}_\textrm{train}$. 
The model is fine-tuned with $\mathcal{T}_\textrm{train}$, and used for the next policy optimization phase.

\subsection{Reduce IL loss weight}

The drawback to using demonstrations to bootstrap policy learning is that those demonstrations may not be optimal with respect to the RL objective, such that the IL loss (\ref{eqn:IL_loss}) could limit the minimum RL loss (\ref{eqn:RL_loss}) that could be achieved without IL.
To account for this possibility, we set $\lambda$ in (\ref{eqn:RLIL_loss}) to decay at each iteration.
By decaying the IL term, we observe the policy initially imitates the demonstrations, then as the algorithm iterations between phases, the same motion style (e.g. alternating tripod-like gaits) qualitatively persist even when as the behavior may otherwise diverge from the demonstrations.

\subsection{Implementation details}

A few additional implementation details helped the algorithm to converge reliably.
Firstly, where the policy in our prior work \cite{whitman2021learning} was not recurrent over time, the policy in this work is recurrent over time. 
We found that a recurrent policy trained more reliably than a non-recurrent policy, due to vanishing gradients over the $T$-step roll-out with the policy and model. 
We apply truncated back-propagation through time \cite{werbos1990backpropagation} to train the recurrent policy to operate for more than $T$ time steps. 
Each time a batch of initial states for the policy roll-out are sampled, half of those states come from the simulation, and half of the initial states come from states approximated by applying the policy to the model. 
The approximated states are associated with recurrent hidden state vectors, which are used as the initial hidden recurrent vector when those approximate states are used as initial states in a subsequent batch.

Secondly, with a fixed or decaying learning rate, the policy optimization would sometimes diverge. 
We used a simple adaptive learning rate rule to stabilize policy optimization.
At the start of each policy optimization phase, we compute the net reward from applying the policy for $T$ steps to a large set (e.g. 10x the batch size) of initial states. 
This ``validation reward'' acts similarly to a validation loss in a supervised learning problem. 
If the current validation reward is lower than the initial validation reward, the policy parameters revert to the last point when the validation reward was computed, the learning rate reduces, and continues.

Occasionally, we encountered a poor-quality local minimum that can occur when the RL and IL loss conflict during early iterations. 
The likelihood of the policy imitating the demonstrations could be low and have a large variance, while the reward obtained is high. 
To prevent this situation, we penalize the entropy of the actions output by the policy when given observations from the demonstration dataset, that is, an additional loss term of the form $\sum_{o \in \Tau_{\textrm{demo}}} H(\pi_\theta(a|o)$ where $H(\cdot)$ is the entropy function. 
This term effectively rewards the policy for being ``certain'' (low variance) about the actions output in the IL loss calculation.
We found this additional loss term stabilized the initial RL+IL iterations.

\begin{algorithm}[tb] 
 \caption{MBRL with IL for modular robots.   
 }
 \label{algo:mbrl_il}
 \begin{algorithmic}[1]

\STATE Train policy $\pi_\theta$ with behavioral cloning from $\mathcal{T}_\textrm{demo}$

  \STATE Collect dataset $\mathcal{T}_\textrm{train}$ from random action trajectories
  \FOR {$i = 1 \dots $ Num. iterations }
   \vspace{0.5em}\STATE  \textit{Model learning phase:}
  \STATE Train model $\tilde f_\phi$ from $\mathcal{T}_\textrm{train}$ with behavioral cloning
    \vspace{0.5em}\STATE \textit{Policy optimization phase:}
    \STATE Initialize buffer $\mathcal{B}$ with random initial robot states and zero-valued policy hidden-state vectors.
      \FOR {$k = 1 \dots K$}
        \STATE Sample a batch of initial states $s_0$ with hidden-state vectors from $\mathcal{B}$, and set hidden state of $\pi_\theta$
        \STATE $R=0$, $\tilde s_0 = s_0$
        \vspace{0.5em}\STATE \textit{Roll out policy with approx. dynamics}
        \FOR{$t = 1 \dots T$ }
            \STATE $a_t \sim \pi_\theta(a_t|O(\tilde s_t))$, $\tilde s_{t+1} = \tilde f_\phi(\tilde s_t, a_t)$
            \STATE $R = R+ r(\tilde s_t, a_t)$
            \ENDFOR
        \STATE $\mathcal{L}_{\textrm{RL}} = -R$
        \IF{ IL weight $\lambda>0$}
            \STATE Sample a batch of $(o, a)$ from $\mathcal{T}_{demo}$
            \STATE Compute $\mathcal{L}_{\textrm{IL}}$ per (\ref{eqn:IL_loss}), $\mathcal{L} = \mathcal{L}_{\textrm{RL}} + \lambda \mathcal{L}_{\textrm{IL}}$
        \ENDIF
        \STATE Update $\pi_\theta$ via gradient descent with $d\mathcal{L}/d\theta$
        \IF{$k \% \textrm{update rate} == 0$}
            \STATE Create a batch of new random initial states, and apply the policy for $T/2$ steps
            \STATE Overwrite part of $\mathcal{B}$ with intermediate states and policy hidden-state vectors, i.e., $\mathcal{B}^{N_\textrm{batch}/2 : N_\textrm{batch}} \leftarrow \{ s_{T/2,i}, h_{T/2} \}_{i=N_\textrm{batch}/2}^{N_\textrm{batch}}$
        \ENDIF
      \ENDFOR

    \vspace{0.5em}\STATE \textit{Data collection:}
    \STATE Apply $\pi_\theta$ to collect data $\mathcal{T}_{\textrm{new}}$
    \STATE $\mathcal{T}_{\textrm{train}}  \leftarrow \mathcal{T}_{\textrm{train}} \cup \mathcal{T}_{\textrm{new}} $
    \STATE Decrease IL weight $\lambda$
  \ENDFOR
 \end{algorithmic} 
 \end{algorithm}

\begin{figure}[!tb] 
\vspace{0.5em}
     \centering
      \includegraphics[width=.95\linewidth]{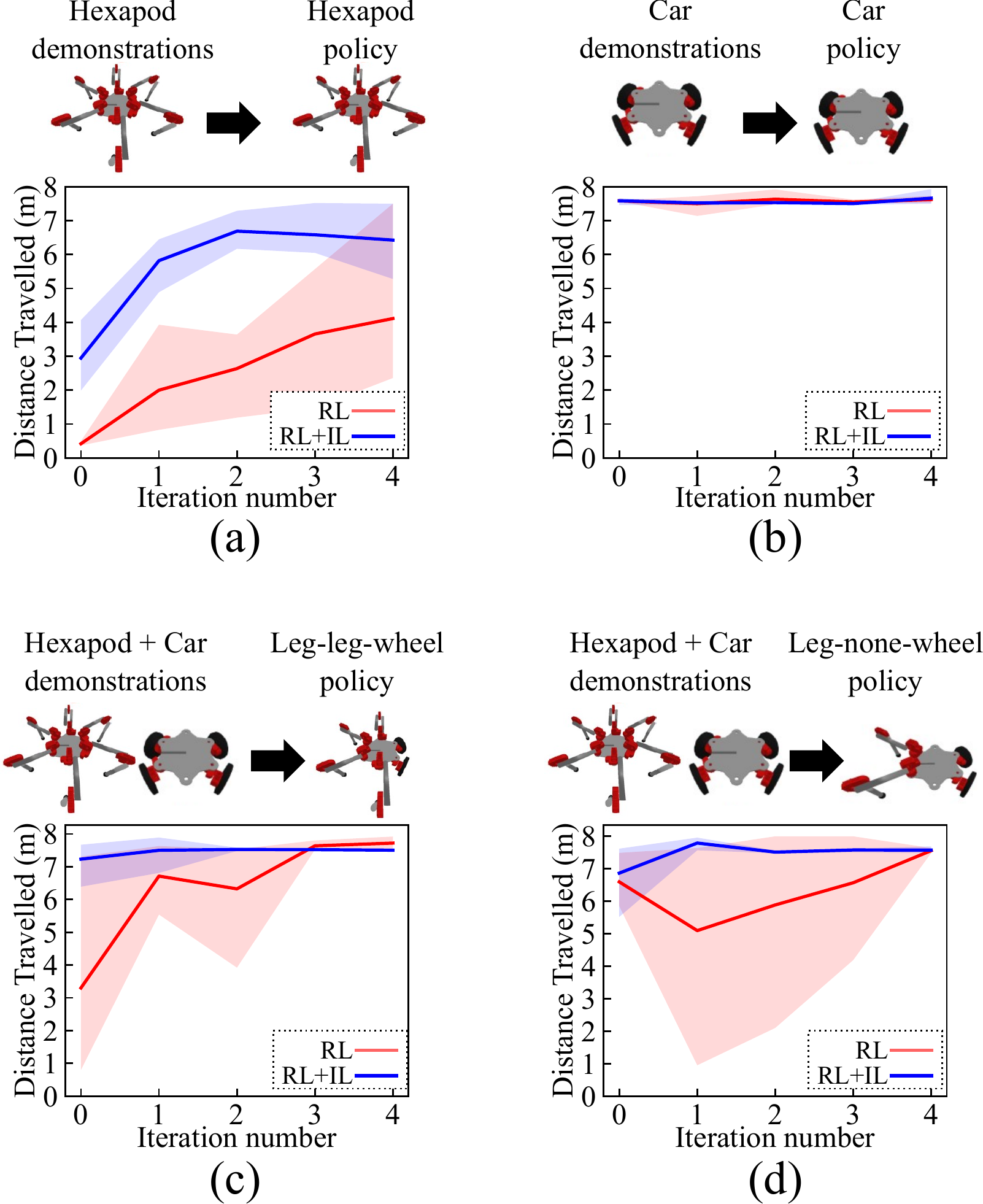}  
  \caption{
   We test our RL+IL method with experiments in which we train robots with different combinations of legs and wheels to locomote.
   Demonstrations come from car-like and hexapod robots (left column), and the policy is trained for robots with either the same or different designs (right column). 
    These plots show the results of experiments described in Sec. \ref{sec:rl_il_experiments}, and compare the outcome of training with and without demonstration data (RL vs. RL+IL). In each plot, the blue curve corresponds to the average over three trials when training with RL+IL method, and the red curve to the RL method without IL. The shaded regions indicate the maximum and minimum value for the trials. (a) A hexapod trained using demonstrations of a hexapod walking with an alternating tripod gait, (b) A car trained using demonstrations of a car driving with skid steering, (c) a robot with four legs and two wheels (leg-leg-wheel) trained using demonstrations from both a hexapod and a car, and (d) a robot with two legs and two wheels (leg-none-wheel) trained using demonstrations from both a hexapod and a car.
    RL+IL required fewer iterations than RL to converge in (a) from unconverged to two iterations, (c) from three to two iterations, and (d) from four to one iteration. RL+IL decreased the variance in convergence rate in all four cases. 
  } \label{fig:rl_il_experiments}
\end{figure}

\section{Experiments}\label{sec:rl_il_experiments}

We conducted experiments to show the benefit of imitating demonstrations from modular designs using our RL+IL algorithm.
The demonstration trajectories $\mathcal{T}_{demo}$ can be derived from robots with either the same designs, or different designs, as the designs the policy is trained to control, as long as all designs involved are made up of different combinations of a set of modules.
We hypothesize that the inclusion of demonstration trajectories can improve policy convergence speed.
We further hypothesize that the improvement in convergence speed will be more noticeable when the training designs are the same as the demonstration designs, but that there will still be an effect when the training and demonstration designs are disjoint sets, i.e., when the design the policy learns to control are not shown by the demonstration dataset.

In these experiments, we use a module set with legs, wheels, and a body.
We create demonstration data for two designs: the four-wheel car (eight D.o.F.) and the six-leg hexapod (eighteen D.o.F.) by hand-crafting behaviors based on expert knowledge.
The hexapod demonstrations show alternating tripod gait trajectories, where the left and right side step sizes are modulated to steer the body yaw towards zero while walking forward.
The car demonstrations show differential drive skid-steering, in which the left and right side wheel velocities are modulated to steer the body yaw towards zero while driving forward.
For both the car and hexapod, we start the robot in 50 different randomly perturbed initial conditions, and create trajectories of 100 time steps showing the robot turning toward zero yaw and moving forward. These trajectories form the dataset $\mathcal{T}_{\textrm{demo}}.$
We used the NVIDIA IsaacGym simulator \cite{makoviychuk2021isaac} with a time step of $1/60$ s, and each action is applied for five time steps.

We test two cases:
firstly, where demonstrations are from the training designs ($D_\textrm{demo} = \{d_\textrm{train} \}$), and secondly, where demonstrations are from different designs ($d_{\text{train}} \notin D_{\text{demo}}$).
In each test, we run the algorithm three times for five iterations, average the results, then plot the max, min, and mean of the distance travelled in 16.7 s (200 simulation time steps).
Each condition trained for about 2 hours on a computer with an 8-core AMD Ryzen 7 2700X processor and a NVIDIA GTX 1070 GPU.
The results of these tests are shown in Fig. \ref{fig:rl_il_experiments}.

In the first two tests, we set $d_\textrm{train}$ to be only either the hexapod (six-leg) or car (four-wheel) design, and measure the effect of including demonstrations in training. 
We found that including hexapod demonstrations improved the convergence speed for the hexapod.
However, since the car is able to learn to locomote nearly optimally in a single iteration, the car demonstrations had little effect on the car policy convergence. 
This result suggests that when the learning task is less complex (lower dimensional, smoother), the demonstrations have less of an effect, but do not impede the RL+IL algorithm.

In the second two tests, we set $d_\textrm{train}$ to be a design with four legs and two wheels, then a design with two legs and two wheels, and measure the effect of including both hexapod and car demonstrations in training. 
The inclusion of demonstrations increased the speed of convergence and decreases the variance, even when the designs learning to locomote are different than the designs shown in the dataset.
The supplementary video shows robots in simulation using the policy trained with and without demonstrations.

\section{Conclusions}

We introduced a novel reinforcement and imitation learning paradigm, where different designs can learn from each other using modular policies. 
This method allows demonstrations from one set of designs can act as a sort of ``prior'' over behaviors for different designs.
Our experiments show that including demonstrations from modular designs can accelerate learning, even when those demonstrations are not from the same robot designs that are being trained, i.e., given a policy for a robot with legs, other robots with a different number of legs can learn from the first robot.

These methods come with some limitations. 
Firstly, the robot design $d_\text{train}$ and those in $D_{\text{demo}}$ must all be made up from different combinations of a set of modules. At present, the method cannot use arbitrary robots, or for instance, demonstrations recorded directly from a human body. 
Secondly, we have not yet validated the output policies on real robots, though our past work \cite{whitman2021learning}  demonstrated that modular policies can control real robots.
Lastly, we do not yet have a baseline for how well the policy performs relative to other policy optimization methods. 
To the best of our knowledge, our method is the first to train one design using data from multiple different designs, so we primarily focused on comparing our method with and without IL.
This comparison is warranted because the algorithm both with and without IL converges to the same average distance travelled, although for the hexapod, the version without IL requires more than the five iterations plotted to reach its full performance level. 
Comparisons to other methods such as MFRL, for instance adapting the methods of \cite{huang2020smp} to include demonstrations, could help establish how well our policies perform relative to policies trained with other methods.

One direction for future work would be to apply RL+IL to train the policy for multiple designs at once.
The formulation and algorithm used here can be readily adapted to include training with a set of multiple designs at once in the RL loss.
Our method would allow for arbitrary combinations of training and demonstration designs.
The RL+IL methods we developed could also be combined with other learning objectives, such as learning to climb using vision, as is the subject of our concurrent work. 
The policy learned on flat ground could be used to warm-start a climbing policy. 
Alternatively, demonstrations from experts could show the robot how to adapt its behavior to different environments.

Future work can also develop additional baseline comparisons, other network architectures, designs, and environments.
While we have focused on a narrow class of locomoting robots in this work, we hope our methods will prove general enough to be applied to different types of robots and settings.
For robots to become more versatile and capable in the real world, we believe they could benefit from learning from other robots that have different designs.


\bibliographystyle{IEEEtran}
\bibliography{main}

\end{document}